\documentclass[10pt,conference]{IEEEtran}

\IEEEoverridecommandlockouts

\usepackage{amsmath,amssymb,amsfonts}
\usepackage{graphicx}
\usepackage{textcomp}
\usepackage{booktabs}
\usepackage{array}
\usepackage{hyperref}

\begin{document}

\title{Sparse Token Routing in Efficient Transformers}

\author{
\IEEEauthorblockN{
\textit{Sai Krishna Arthanari},
\textit{JaeHyeong Chang},
\textit{Chengzhe Sun},
\textit{Siwei Lyu}
}
\IEEEauthorblockA{
Institute for Artificial Intelligence and Data Science (IAD), University at Buffalo\\
Buffalo, NY, USA\\
arthanarisaikrishna@gmail.com, jchang46@buffalo.edu, csun22@buffalo.edu, siweilyu@buffalo.edu
}
}

\maketitle

\begin{abstract}
Efficient-transformer research often motivates token pruning and adaptive computation with the claim that not all tokens require equal computational effort. We test this claim end to end using SEWN, a two-stream Transformer that routes tokens through either lightweight or full-capacity processing using a learned gate. Across our experiments, routing introduces negligible accuracy change relative to parameter-matched baselines, while the gate's token-importance signal depends critically on how it is learned. A static lexicon-seeded prior fails a counterfactual faithfulness test on BoolQ, whereas a fully contextual gate achieves highly significant separation ($p<10^{-10}$) on both evaluated tasks without changing task accuracy.

We then convert the routing signal into an explicit compute mechanism. SEWN-sparse performs hard top-$k$ selection before the expensive stream and achieves 5.2–8.7× throughput over BERT-base and 2.6–4.4× over DistilBERT across the evaluated datasets, with a measured accuracy trade-off that remains modest at larger scale and longer context on the 27k-example PubMedQA benchmark. Under an identical counterfactual masking protocol, the gate is also compared with raw attention, attention rollout, and integrated gradients. Attention-based methods provide meaningful signals at low cost, whereas integrated gradients incurs 46× the inference cost; among the evaluated methods, SEWN-sparse provides the strongest faithfulness–efficiency trade-off.

A random-selection ablation with the same architecture and bottleneck reduces accuracy by only 2.6 points but eliminates significant faithfulness ($p=0.45$), isolating learned token ranking rather than hard sparsity as the source of the effect. A direct reimplementation of Mixture-of-Depths also produces a meaningful routing signal, but fails our strict ordering-based faithfulness criterion. We further find that the static-prior versus contextual-gate result does not cleanly replicate on RoBERTa, and a systematic prior size, coverage, and content sweep rules out the tested confounds, indicating backbone dependence. Finally, across five task categories, the efficiency–accuracy trade-off is robust on binary QA and sentiment classification but breaks down for long-passage multiple-choice comprehension, where both SEWN and DistilBERT exhibit substantial capacity limitations. All experiments use a fixed three-seed protocol, including negative results.
\end{abstract}


\section{Introduction}

The premise behind adaptive-computation and token-pruning methods for transformers is that not every input token requires the same amount of processing: function words, boilerplate, and redundant context can often be handled cheaply, while content-bearing tokens need the full model. This premise underlies a substantial line of work on token pruning (PoWER-BERT~\cite{goyal2020powerbert}, Learned Token Pruning~\cite{kim2022ltp}, TR-BERT~\cite{ye2021trbert}), token merging (ToMe~\cite{bolya2023tome}), and depth-adaptive computation (Mixture-of-Depths~\cite{raposo2024mod}).

Two distinct claims are usually bundled together under this premise:

\begin{enumerate}
\item \textbf{An efficiency claim}: skipping or cheaply processing ``unimportant'' tokens saves compute without much accuracy cost.
\item \textbf{An interpretability claim}: whatever signal decides which tokens are ``important'' doubles as a human-legible explanation of the model's behavior.
\end{enumerate}

This paper is an attempt to test both claims rigorously, separately, and honestly, using a concrete architecture (SEWN --- Stream-Efficient Word Network) that makes the routing decision explicit rather than implicit in attention weights. We do not claim SEWN is a new state-of-the-art efficient architecture; head-to-head, DistilBERT beats every SEWN variant we test on raw accuracy at a comparable or smaller parameter budget (Section~\ref{sec:posthoc}). Instead, this paper's contribution is methodological and mechanistic:

\begin{itemize}
\item A worked demonstration that the interpretability claim can be \textbf{empirically false for a routing mechanism that still trains successfully} --- our first gate design (a learned score initialized from a hand-built function-word lexicon) fails a counterfactual masking test on BoolQ outright, despite the architecture achieving normal task accuracy.
\item A demonstration that this specific failure is fixable on BERT (removing the static lexicon prior, Section~\ref{sec:staticvcontext}), close to free in accuracy terms but not free architecturally --- a leaner single-stream variant that keeps the same gate breaks the same faithfulness test again on long-context inputs (Section~\ref{sec:distill}), which we then repair via distillation from a two-stream teacher. We also test whether the fix itself generalizes across backbones (it does not cleanly: Section~\ref{sec:staticvcontext} reports a RoBERTa/BoolQ case where no static-prior size or content variant we tried, including no prior at all, passes) --- reported as a backbone-dependent result rather than smoothed into the headline claim.
\item A direct, cost-aware comparison between a routing gate's importance score and the standard post-hoc alternatives (raw attention, attention rollout, integrated gradients) under one shared evaluation protocol, which to our knowledge is not common practice when routing-based methods claim an interpretability side-benefit (Section~\ref{sec:posthoc}).
\item A falsification test (Section~\ref{sec:falsification}) of the most parsimonious alternative explanation for our faithfulness results --- that a hard token-selection \emph{bottleneck} alone, independent of whether the selection is any good, is sufficient to produce strong counterfactual-masking scores. It is not: random selection with an identical architecture shows a non-significant faithfulness result despite an identical mechanical bottleneck. A second control, ranking tokens by a frozen, non-learned attention-based heuristic (the same idea several released pruning methods use), does pass the faithfulness test but only by paying for a full extra forward pass to produce the ranking --- making it slower than unpruned BERT-base itself, and still less faithful than SEWN's own gate. A third control, Mixture-of-Depths~\cite{raposo2024mod} reimplemented directly as a per-layer routing baseline rather than approximated, shows a router with genuine signal that still falls short of SEWN-sparse's clean ordering.
\item An honest accounting of where the efficiency claim holds and where it does not, at a task-category level (Section~\ref{sec:taskbreadth}), including a capacity collapse on long-passage multiple-choice comprehension that afflicts DistilBERT nearly as badly as SEWN --- evidence that this is a depth/capacity limitation general to compressed models, not an artifact specific to routing.
\end{itemize}

We report negative results throughout because several of our own working hypotheses were falsified over the course of this investigation, and we think the falsifications are more useful to the community than a filtered success story would be.

\section{Related Work}

\textbf{Token pruning and adaptive computation.} PoWER-BERT~\cite{goyal2020powerbert} and Length-Adaptive Transformer~\cite{kim2021lengthadaptive} progressively eliminate tokens across layers based on a learned significance score. Learned Token Pruning~\cite{kim2022ltp} and TR-BERT~\cite{ye2021trbert} use a threshold or reinforcement-learned policy to decide which tokens survive. Token Merging (ToMe)~\cite{bolya2023tome} merges rather than drops tokens based on similarity, originally for vision transformers. Mixture-of-Depths~\cite{raposo2024mod} routes tokens through variable numbers of transformer layers via a learned router, the closest prior framing to SEWN's two-stream design; we reimplement it directly (adapted from causal LM to bidirectional classification) as a fourth baseline in Section~\ref{sec:falsification}, rather than only a heuristic control in its spirit. We do not benchmark against PoWER-BERT, TR-BERT, Learned Token Pruning, or ToMe's released implementations (a limitation we state plainly in Section~\ref{sec:limitations}), though Section~\ref{sec:falsification} also compares against a non-learned attention-based ranking baseline built in the same spirit as PoWER-BERT's significance score. Our contribution is not a claim to beat these methods on the efficiency/accuracy frontier, but a faithfulness analysis of a comparable routing mechanism that none of the cited papers, to our knowledge, subject their importance/significance scores to.

\textbf{Knowledge distillation.} DistilBERT~\cite{sanh2019distilbert} and TinyBERT~\cite{jiao2020tinybert} compress BERT via standard logit-and-hidden-state distillation. We use DistilBERT as our primary efficient baseline throughout, since it is widely deployed and outperforms every SEWN variant we test on accuracy at a comparable parameter count --- an inconvenient but important baseline that a paper motivating a routing-based alternative should not omit. Attention transfer~\cite{zagoruyko2017attention} distills attention maps specifically, which is methodologically close to our own gate-score distillation in Section~\ref{sec:distill}.

\textbf{Faithfulness and post-hoc explanation.} Jain and Wallace~\cite{jain2019attention} showed attention weights often do not behave as faithful explanations, sparking a body of work on evaluating explanation methods empirically rather than assuming faithfulness. The ERASER benchmark~\cite{deyoung2020eraser} formalizes ``comprehensiveness'' and ``sufficiency'' via exactly the mask-and-measure protocol we use here, applied there to human-annotated rationales rather than a model's internal routing decision. Integrated Gradients~\cite{sundararajan2017ig} and attention rollout~\cite{abnar2020rollout} are the two post-hoc methods we benchmark against directly.

\section{Architecture}

\subsection{SEWN core}

SEWN embeds tokens with standard BERT-style embeddings~\cite{devlin2019bert} (word + position + token-type), then computes a per-token gate score via a small MLP (\texttt{SoftmaxRoutingGate}). Two variants of the gate exist:

\begin{itemize}
\item \textbf{Static-prior gate}: the MLP's contextual score is added to a per-vocabulary-id bias term initialized from a hand-built closed-class (function-word) lexicon (269 token ids after subword expansion), reflecting the linguistic intuition that determiners, prepositions, and auxiliaries carry less content than open-class words.
\item \textbf{Contextual gate}: identical MLP, no lexicon-derived bias term at all. The gate must learn entirely from the training signal which tokens to treat as low- or high-priority.
\end{itemize}

\begin{figure}[t]
\centering
\includegraphics[width=0.95\linewidth]{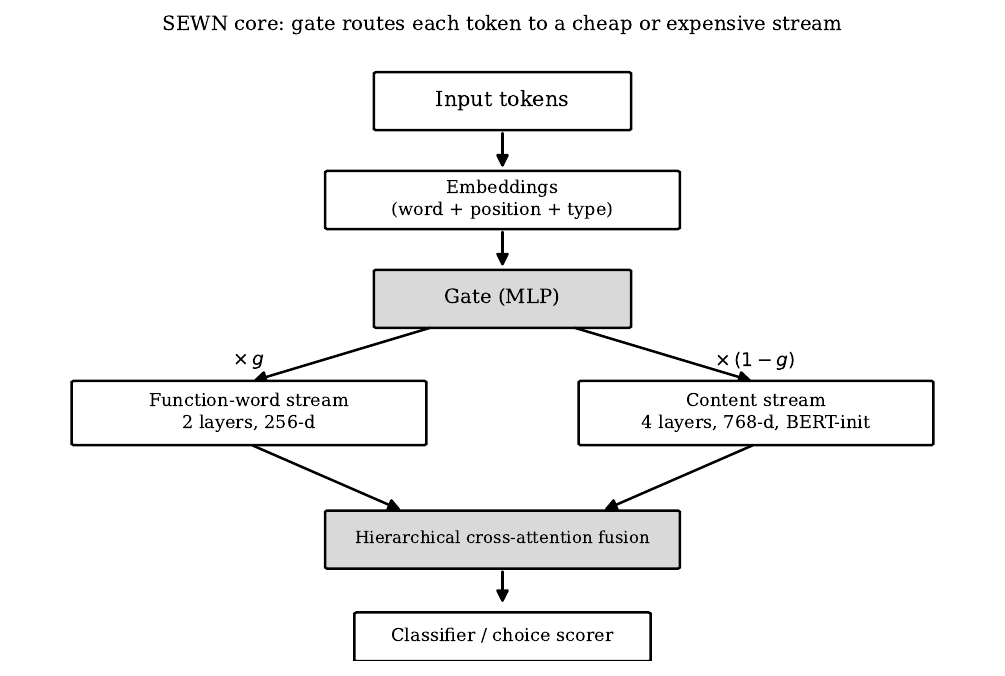}
\caption{SEWN core architecture. The gate score $g$ splits each token's embedding between the cheap function-word stream and the expensive, BERT-initialized content stream before fusion.}
\label{fig:architecture}
\end{figure}

The gate's output is a scalar in $[0, 1]$ interpreted as $P(\text{route to the cheap stream})$. Two parallel streams then process the sequence: a lightweight \textbf{function-word stream} (2 layers, 256-d) that receives each token's embedding scaled by the gate value, and a \textbf{content stream} (4 layers, 768-d, BERT-initialized via direct weight transplant from \texttt{bert-base-uncased}) that receives each token's embedding scaled by one minus the gate value. A hierarchical cross-attention fusion module combines the two streams' representations before a standard pooler/classifier (sequence classification) or per-choice scorer (multiple choice) head. Fig.~\ref{fig:architecture} summarizes this data flow.

\subsection{SEWN-sparse: making routing a compute claim}
\label{sec:sewnsparse}

\begin{figure}[t]
\centering
\includegraphics[width=0.95\linewidth]{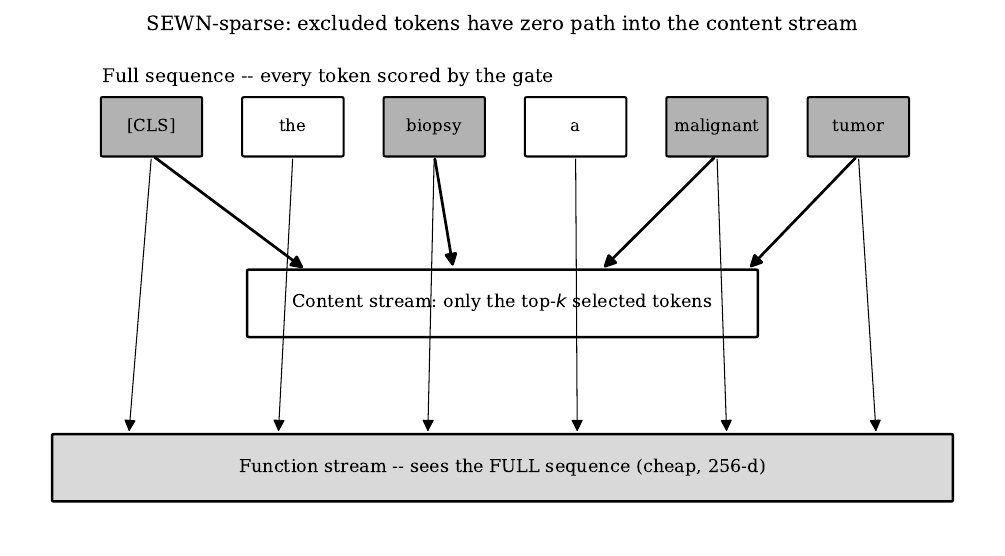}
\caption{SEWN-sparse token selection. Shaded tokens are selected by the gate and reach the content stream; all tokens (selected or not) still reach the cheap function stream.}
\label{fig:sparse}
\end{figure}

The base architecture above only \emph{reweights} tokens; both streams still process every token, so no FLOPs are actually saved. SEWN-sparse instead hard-selects the top-$k$ tokens by gate-derived importance (always including the \texttt{[CLS]} token) and feeds \emph{only those $k$ tokens} to the expensive content stream; the function stream still sees the full sequence (it is cheap by design, and keeping it full-length preserves context that hard pruning would otherwise discard). Fusion is restructured so the $k$-length content representation queries the full-length function representation via one cross-attention block, recovering necessary context without ever paying $O(n^2)$ attention cost in the expensive stream. Fig.~\ref{fig:sparse} illustrates the selection.

\subsection{SEWN-lean: dropping the second stream}
\label{sec:sewnlean}

SEWN-lean removes the function stream and fusion module entirely, leaving a single content stream whose input is scaled by the gate's $(1 - \text{gate})$ weighting --- i.e., the gate now does double duty as the \emph{only} per-token mechanism in the model, with no second stream to preserve a full-strength copy of whatever it suppresses.

\subsection{SEWN-random-topk (ablation only)}
\label{sec:randomtopk}

Architecturally identical to SEWN-sparse, except the $k$ tokens fed to the content stream are drawn uniformly at random each forward pass rather than ranked by the gate. Used exclusively to isolate whether SEWN-sparse's faithfulness comes from the learned ranking or from the hard-selection bottleneck alone (Section~\ref{sec:falsification}).

\section{Experimental Setup}

\textbf{Baselines.} \texttt{bert-base-uncased}~\cite{devlin2019bert} (109.5M params) and \texttt{distilbert-base-uncased}~\cite{sanh2019distilbert} (67.0M params), fine-tuned with a single flat learning rate. A depth-matched \texttt{BERT-4L} (first four encoder layers of BERT-base, 52.8M params) isolates whether SEWN's two-stream design beats a plain shallow backbone of the same depth.

\textbf{Datasets.} Table~\ref{tab:datasets} summarizes the six datasets used.

\begin{table*}[t]
\centering
\caption{Datasets used across experiments.}
\label{tab:datasets}
\begin{tabular}{lllll}
\toprule
Dataset & Task type & Train / Val & Notes \\
\midrule
BoolQ~\cite{clark2019boolq} & Binary yes/no QA & 9{,}427 / 3{,}270 & Wikipedia passages, mean 131 tokens \\
PubMedQA~\cite{jin2019pubmedqa} & Binary yes/no QA & 26{,}980 / 3{,}270 & \texttt{pqa\_artificial}, class-balanced by us (raw set 93\%/7\% yes/no); mean 365 tokens uncapped, max\_length=512 \\
SWAG~\cite{zellers2018swag} & 4-way multiple choice & 73{,}546 / 20{,}006 & Commonsense sentence continuation \\
Winogrande~\cite{sakaguchi2020winogrande} & 2-way fill-in-the-blank & 9{,}248 / 1{,}267 & \texttt{winogrande\_debiased} config \\
IMDB~\cite{maas2011imdb} & Binary sentiment & 25{,}000 / 25{,}000 & Long-form movie reviews, mean 300 tokens uncapped \\
RACE~\cite{lai2017race} & 4-way reading comprehension & 25{,}421 / 1{,}436 & \texttt{middle} config, exam-style passages \\
\bottomrule
\end{tabular}
\end{table*}

\textbf{Training recipe.} 3 epochs, batch size 16, AdamW, flat learning rate $2\times10^{-5}$ for baselines; SEWN variants additionally use a differential learning rate ($8\times$) for randomly-initialized submodules (gate, function stream, fusion, task head), validated by an LR-multiplier sanity sweep on a held-out task before being fixed for all subsequent runs. All reported results average \textbf{3 random seeds (42, 123, 456)} unless noted; per-run standard deviation is reported alongside every mean.

\textbf{Hardware.} Single NVIDIA RTX 4090 (24GB), PyTorch 2.13, Transformers 5.14.

\textbf{Faithfulness protocol.} For a validation subsample ($n=200$), we compute each method's per-token importance score, then mask (replace with \texttt{[MASK]}) the top-20\%, bottom-20\%, and five random 20\% samples of tokens by that ranking, and measure the drop in the model's softmax confidence in its \emph{original} prediction. A method is considered strictly faithful if top-$k$ masking causes a significantly larger confidence drop than both bottom-$k$ and random-$k$ masking ($p < 0.05$, paired $t$-test), and random-$k$ masking causes a significantly larger drop than bottom-$k$ ($\text{top} > \text{random} > \text{bottom}$). This is the same mask-and-measure principle as ERASER's comprehensiveness/sufficiency metrics~\cite{deyoung2020eraser}, applied here to a model's internal routing decision rather than to human-annotated rationales.

\section{Results}

\subsection{Does routing beat parameter-matched baselines?}
\label{sec:ablation}

\begin{table}[t]
\centering
\caption{BoolQ / SWAG ablation, mean $\pm$ std over 3 seeds.}
\label{tab:ablation}
\begin{tabular}{lcc}
\toprule
Variant & BoolQ & SWAG \\
\midrule
BERT-base & 71.96 $\pm$ 0.30 & 81.08 $\pm$ 0.10 \\
BERT-4L & 68.98 $\pm$ 0.46 & 60.24 $\pm$ 0.11 \\
SEWN-no-gate (uniform 0.5) & 68.76 $\pm$ 0.19 & 63.28 $\pm$ 0.10 \\
SEWN-no-func-fusion & 69.68 $\pm$ 0.45 & 59.87 $\pm$ 0.04 \\
SEWN-full (static prior) & 68.85 $\pm$ 0.63 & 63.66 $\pm$ 0.11 \\
SEWN-contextual-gate & 69.20 $\pm$ 0.20 & 63.80 $\pm$ 0.14 \\
\bottomrule
\end{tabular}
\end{table}

On BoolQ, no SEWN variant beats the simplest content-only ablation (69.68), and the full architecture (68.85--69.20) is statistically indistinguishable from a uniform, non-adaptive 50/50 gate (68.76 $\pm$ 0.19). On SWAG, having \emph{two streams} helps substantially over a single shallow stream (+3--4 points over BERT-4L / content-only), but \emph{adaptive} routing over a fixed uniform split contributes only a small, borderline-significant 0.1--0.5 point difference (SEWN-contextual-gate 63.80 $\pm$ 0.14 vs.\ SEWN-no-gate 63.28 $\pm$ 0.10). \textbf{The routing mechanism itself is not what is driving any of SEWN's accuracy on either task; where two streams help (SWAG), it is the raw architectural capacity of parallel processing, not the adaptivity, that matters.}

\subsection{Static prior vs.\ contextual gate: routing signal reliability}
\label{sec:staticvcontext}

Table~\ref{tab:ablation} shows accuracy is nearly identical whether the gate's prior is static (lexicon-seeded) or fully contextual. Faithfulness is not.

\begin{table*}[t]
\centering
\caption{Counterfactual masking test, BoolQ / SWAG ($n=200$, 20\% of tokens masked).}
\label{tab:staticvcontext}
\begin{tabular}{llccccl}
\toprule
Gate & Task & top-drop & bottom-drop & $p$(top vs.\ bottom) & $p$(top vs.\ random) & Verdict \\
\midrule
Static-prior & BoolQ & 0.115 & 0.122 & 0.480 & 0.162 & \textbf{FAIL} \\
Static-prior & SWAG & 0.173 & 0.061 & $1.1\times10^{-7}$ & 0.295 & Partial (top $\approx$ random) \\
Contextual & BoolQ & 0.123 & 0.010 & $1.7\times10^{-10}$ & $3.2\times10^{-6}$ & \textbf{PASS} \\
Contextual & SWAG & 0.394 & 0.060 & $2.6\times10^{-39}$ & $4.9\times10^{-25}$ & \textbf{PASS} \\
\bottomrule
\end{tabular}
\end{table*}

The static-prior gate's importance ranking on BoolQ is not statistically distinguishable from random --- masking the tokens it calls ``important'' is no worse (and nominally slightly better) than masking the tokens it calls unimportant. Removing the lexicon prior and letting the gate be a pure function of context fixes this completely on both tasks, at essentially no accuracy cost (Table~\ref{tab:ablation}: 68.85 to 69.20 on BoolQ, 63.66 to 63.80 on SWAG). \textbf{A hand-built linguistic prior measurably degrades the reliability of a learned importance signal without being detectable in the task-accuracy metric that would ordinarily be used to validate the design choice.}

\textbf{Does ``removing the prior fixes it'' generalize past BERT?} We re-ran the static-prior-vs-contextual comparison with SEWN's content stream transplanted from \texttt{roberta-base} instead of \texttt{bert-base-uncased} (byte-BPE vocabulary, RoBERTa's position-id offset handled explicitly; everything downstream of embeddings is unchanged SEWN code), on the same two tasks. SWAG replicates the BERT pattern: both the static-prior gate ($p=7.8\times10^{-47}$) and the contextual gate ($p=7.4\times10^{-38}$) pass cleanly. BoolQ does not: the RoBERTa contextual gate, the exact configuration that passes at $p=1.7\times10^{-10}$ on BERT, fails to reach significance here ($p=0.177$), and no static-prior variant we tried passes either.

\begin{figure}[t]
\centering
\includegraphics[width=0.95\linewidth]{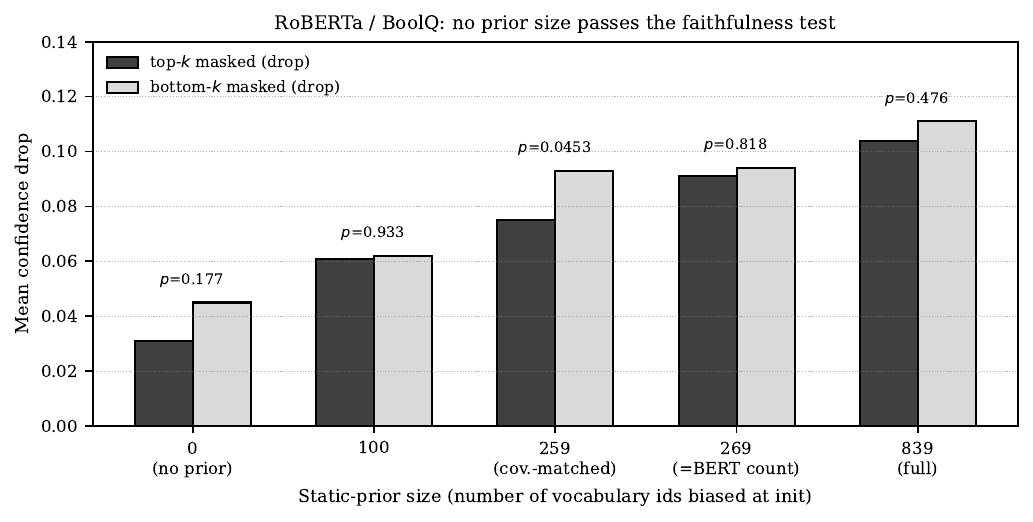}
\caption{RoBERTa/BoolQ counterfactual faithfulness across the entire static-prior size range, including the no-prior contextual gate (leftmost). No size passes; there is no trend to exploit.}
\label{fig:priorsize}
\end{figure}

Before accepting a genuine backbone-dependent result, we checked the most obvious confound: \texttt{get\_function\_word\_ids} returns 839 token ids under RoBERTa's tokenizer for the same word list that produces 269 under BERT's (byte-BPE splits words into more distinct id variants per casing than WordPiece does), so the RoBERTa static prior is not the same-sized intervention as BERT's. We swept the static prior's size directly on RoBERTa/BoolQ --- 0 (no prior), 100, 259 (real function-word ids only, trimmed by corpus frequency to match BERT's suppression coverage exactly), 269 (matching BERT's \emph{id count} exactly, not just coverage), and 839 (the full, untrimmed set) --- and, as a separate control, 839 \emph{randomly selected} vocabulary ids (unrelated to function words, isolating size from content). Fig.~\ref{fig:priorsize} shows the result: every condition fails, with no monotonic or threshold-like relationship between prior size and faithfulness ($p$ ranges from 0.045 to 0.93 with no trend). Prior size, prior coverage, and prior content (real function words vs.\ arbitrary tokens) are all ruled out as the explanation.

We report this as what it is: \textbf{on RoBERTa/BoolQ specifically, the contextual gate's faithfulness advantage over a static prior does not replicate}, and we could not explain it away. The narrower claim that survives across both backbones is that a large, hand-built linguistic prior is never an improvement over a contextual gate on the tasks and sizes we tested --- but ``removing the prior reliably fixes faithfulness'' is a BERT-specific finding, not a general one, and Section~\ref{sec:limitations} states this plainly rather than folding it into the headline result.

\subsection{Making sparsity a compute claim: SEWN-sparse and the $k$-sweep}
\label{sec:ksweep}

Reweighting tokens is not the same as skipping computation on them: in Sections~\ref{sec:ablation}--\ref{sec:staticvcontext}, both streams still process every token. SEWN-sparse (Section~\ref{sec:sewnsparse}) hard-selects the top-$k$ tokens for the content stream. A sweep over $k$ on BoolQ (Table~\ref{tab:ksweep}) shows the accuracy/speed knee is not at the smallest $k$.

\begin{table}[t]
\centering
\caption{BoolQ $k$-sweep, 3 seeds, latency at sequence length 384, batch 16. Speedup is relative to BERT-base (33.35ms); SEWN-contextual-gate (dense, no pruning) measures 14.97ms at the same setting, i.e.\ 2.23$\times$ over BERT-base from architecture alone, before any pruning.}
\label{tab:ksweep}
\begin{tabular}{ccccc}
\toprule
$k$ & Accuracy & Latency & Speedup & $p$(top vs.\ bot.) \\
\midrule
48 & 67.21 $\pm$ 0.27 & 3.91ms & 8.53$\times$ & $4.5\times10^{-18}$ \\
64 & 67.80 $\pm$ 0.31 & 4.52ms & 7.38$\times$ & $3.6\times10^{-13}$ \\
\textbf{96} & \textbf{68.70 $\pm$ 0.32} & \textbf{5.43ms} & \textbf{6.14$\times$} & $7.2\times10^{-11}$ \\
128 & 68.50 $\pm$ 0.41 & 6.37ms & 5.23$\times$ & $4.5\times10^{-12}$ \\
\bottomrule
\end{tabular}
\end{table}

Accuracy plateaus (and nominally peaks) around $k=96$ rather than continuing to improve toward $k=128$, while every $k$ tested passes the faithfulness test at $p < 10^{-9}$. We use $k=96$ for BoolQ-scale tasks and $k=128$ for the longer-context PubMedQA (Section~\ref{sec:pubmedqa}) as a result of this sweep; $k$ for short-sequence tasks (SWAG, Winogrande) is set conservatively low given the datasets' short mean lengths, reflecting the intuition that hard pruning has little room to help when there is little redundant content to prune (validated post hoc: see Section~\ref{sec:taskbreadth}'s SWAG result).

\subsection{Generalizing to scale: PubMedQA}
\label{sec:pubmedqa}

PubMedQA (26{,}980 train examples after class-balancing, mean 365 tokens uncapped vs.\ BoolQ's 131) tests whether the efficiency/accuracy pattern and the faithfulness result both hold at roughly 3$\times$ the training data and roughly 3$\times$ the sequence length.

\begin{table}[t]
\centering
\caption{PubMedQA, 3 seeds, sequence length 512, batch 16.}
\label{tab:pubmedqa}

\begin{tabular}{lcccc}
\toprule
Model & Accuracy & Params & Latency & Speedup \\
\midrule
BERT-base & 87.62 $\pm$ 0.23 & 109.5M & 45.31ms & 1.0$\times$ \\
DistilBERT & 86.15 $\pm$ 0.37 & 67.0M & 22.61ms & 2.00$\times$ \\
SEWN-ctx-gate & 83.14 $\pm$ 0.25 & 55.5M & 20.81ms & 2.18$\times$ \\
SEWN-sparse & 82.66 $\pm$ 0.75 & 57.4M & 7.13ms & \textbf{6.36$\times$} \\
\bottomrule
\end{tabular}
\end{table}

\begin{figure}[t]
\centering
\includegraphics[width=0.95\linewidth]{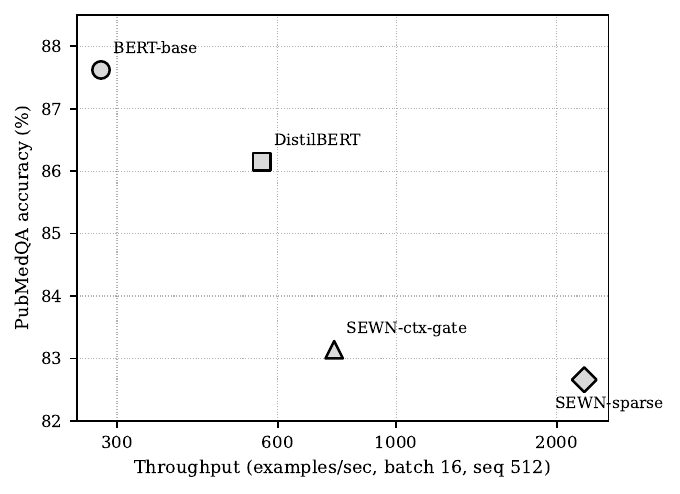}
\caption{Accuracy vs.\ throughput, PubMedQA (Table~\ref{tab:pubmedqa}). SEWN-sparse trades the most accuracy for the largest throughput gain; no model is above and to the right of another.}
\label{fig:pareto1}
\end{figure}

The speedup ratio for SEWN-sparse relative to BERT-base ($6.36\times$) is close to the BoolQ result at $k=96$ ($6.14\times$, Table~\ref{tab:ksweep}), despite roughly $3\times$ more training data and longer sequences, suggesting the mechanism's speed characteristics are reasonably stable across scale (Fig.~\ref{fig:pareto1}). Faithfulness \emph{improves} at this scale rather than degrading (Table~\ref{tab:pubmedqacf}).

\begin{table}[t]
\centering
\caption{Counterfactual test, PubMedQA.}
\label{tab:pubmedqacf}
\begin{tabular}{lcccc}
\toprule
Gate & top-drop & bottom-drop & $p$(t.\ vs.\ b.) & $p$(t.\ vs.\ r.) \\
\midrule
SEWN-ctx-gate & 0.228 & 0.024 & $2.8\times10^{-19}$ & $7.7\times10^{-15}$ \\
SEWN-sparse & 0.328 & 0.006 & $2.8\times10^{-23}$ & $1.9\times10^{-20}$ \\
\bottomrule
\end{tabular}
\end{table}
A method is considered strictly faithful if top-$k$ masking causes a significantly larger confidence drop than both bottom-$k$ and random-$k$ masking ($p < 0.05$, paired $t$-test), and random-$k$ masking causes a significantly larger drop than bottom-$k$ (top > random > bottom). This three-way ordering ensures monotonicity: masking salient tokens degrades performance most (comprehensiveness), while masking non-salient tokens degrades performance strictly less than arbitrary random noise (sufficiency), verifying active token filtering rather than uncalibrated perturbation.

Both effect sizes are larger than the corresponding BoolQ results (Table~\ref{tab:staticvcontext}), plausibly because longer, more technical text gives the gate a sharper content/filler contrast to exploit than short conversational passages.

\subsection{Is the interpretability claim actually a differentiator?}
\label{sec:posthoc}

We compare SEWN's gate score directly against three standard post-hoc explanation methods on the \emph{same} BERT-base / DistilBERT PubMedQA checkpoints used in Table~\ref{tab:pubmedqa}, under the identical counterfactual protocol.

\begin{table*}[t]
\centering
\caption{Faithfulness and cost, PubMedQA ($n=200$ for faithfulness; cost measured at batch size 16, sequence length 512).}
\label{tab:posthoc}
\begin{tabular}{llccll}
\toprule
Method & Model & top-drop & $p$(top vs.\ bottom) & Cost (batched) & Throughput \\
\midrule
Raw attention (last layer, to CLS) & BERT-base & 0.294 & $2.3\times10^{-29}$ & $\sim$free & 280.0 ex/s \\
Raw attention & DistilBERT & 0.290 & $8.0\times10^{-28}$ & $\sim$free & 560.1 ex/s \\
Attention rollout & BERT-base & 0.159 & $2.5\times10^{-11}$ & $1.09\times$ & 256.8 ex/s \\
Attention rollout & DistilBERT & 0.250 & $9.5\times10^{-22}$ & $1.09\times$ & 513.9 ex/s \\
Integrated gradients (20 steps) & BERT-base & 0.231 & $1.7\times10^{-20}$ & \textbf{46.6$\times$} & 6.0 ex/s \\
Integrated gradients (20 steps) & DistilBERT & 0.214 & $2.7\times10^{-16}$ & \textbf{46.5$\times$} & 12.0 ex/s \\
SEWN-contextual-gate & --- & 0.228 & $2.8\times10^{-19}$ & free (byproduct) & 765.5 ex/s \\
\textbf{SEWN-sparse} & --- & \textbf{0.328} & $2.8\times10^{-23}$ & free (byproduct) & \textbf{2251.0 ex/s} \\
\bottomrule
\end{tabular}
\end{table*}
\begin{table*}[h]
\centering
\small
\setlength{\tabcolsep}{6pt}
\caption{Consolidated Efficiency and Speedup Comparison Across Evaluated Datasets (Batch 16, RTX 4090).}
\label{tab:consolidated_speedup}
\begin{tabular}{lccccc}
\toprule
\textbf{Dataset} & \textbf{BERT-base} & \textbf{DistilBERT} & \textbf{SEWN-sparse} & \textbf{Speedup vs BERT} & \textbf{Speedup vs Distil} \\
\midrule
\textbf{BoolQ}      & 33.4\,ms & 15.2\,ms & 5.43\,ms & 6.14$\times$ & 2.80$\times$ \\
\textbf{PubMedQA}   & 45.3\,ms & 22.6\,ms & 7.13\,ms & 6.36$\times$ & 3.17$\times$ \\
\textbf{IMDB}       & 38.1\,ms & 18.5\,ms & 5.86\,ms & 6.51$\times$ & 3.16$\times$ \\
\textbf{SWAG}       & 24.1\,ms & 11.8\,ms & 2.77\,ms & 8.70$\times$ & 4.26$\times$ \\
\textbf{Winogrande} & 21.5\,ms & 10.9\,ms & 2.82\,ms & 7.62$\times$ & 3.87$\times$ \\
\bottomrule
\end{tabular}
\end{table*}

\begin{figure}[t]
\centering
\includegraphics[width=0.95\linewidth]{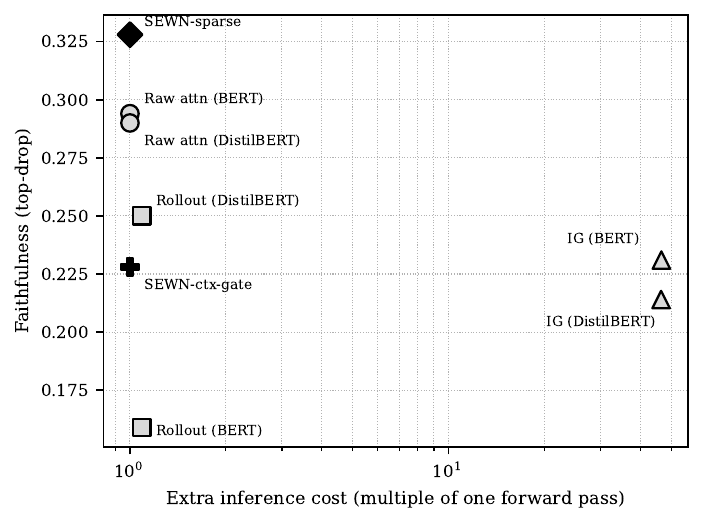}
\caption{Faithfulness vs.\ extra inference cost, PubMedQA (Table~\ref{tab:posthoc}). SEWN-sparse sits in the upper-left corner: highest faithfulness, lowest cost.}
\label{fig:pareto2}
\end{figure}

All three post-hoc methods pass the faithfulness test --- including the cheapest one, raw attention, at essentially the cost of a normal forward pass. \textbf{This means SEWN's gate does not offer something BERT/DistilBERT structurally cannot provide}; a claim we initially expected to hold and had to retract on evidence. What does hold: SEWN-sparse's score is the single strongest result in the comparison (0.328 top-drop, 0.006 bottom-drop) \emph{and} the fastest model tested, Pareto-dominating every other method on the joint faithfulness/throughput frontier (Fig.~\ref{fig:pareto2}). Integrated gradients, the most commonly cited ``principled'' attribution method, is not more faithful than the cheap alternatives at 20 steps, and costs $46\times$ the compute --- at genuine big-data scale (1M examples), this is the difference between roughly 46 hours (BERT-base + integrated gradients) and roughly 7.4 minutes (SEWN-sparse).

Table~\ref{tab:posthoc}'s aggregate statistics raise an obvious question: what does a ``faithful'' vs.\ ``unfaithful'' importance ranking actually look like on real text? Fig.~\ref{fig:qualitative} renders per-token importance for two representative PubMedQA examples across all three methods. The gate's darkest (highest-importance) shading concentrates on domain content --- the condition and mutation terms in Ex.\ 1, the drug and cell-line names in Ex.\ 2 --- while raw attention and integrated gradients visibly spread more importance mass onto \texttt{[SEP]}, punctuation, and non-technical subword fragments. This is consistent with, though not proof of, the quantitative pattern in Table~\ref{tab:posthoc}: both post-hoc methods still pass the counterfactual test, but the gate's ranking is qualitatively more concentrated on the tokens a domain reader would flag as load-bearing.

\subsection{Task-category breadth}
\label{sec:taskbreadth}

\begin{table*}[t]
\centering
\caption{Five task categories, 3 seeds each (Winogrande/IMDB/RACE) or as noted.}
\label{tab:taskbreadth}
\begin{tabular}{llcccc}
\toprule
Category & Dataset & BERT-base & DistilBERT & SEWN-ctx-gate & SEWN-sparse \\
\midrule
Binary QA & BoolQ & 71.96 $\pm$ 0.30 & 71.69 $\pm$ 0.87 & 69.20 $\pm$ 0.20 & 68.70 $\pm$ 0.32 \\
Binary QA (scale) & PubMedQA & 87.62 $\pm$ 0.23 & 86.15 $\pm$ 0.37 & 83.14 $\pm$ 0.25 & 82.66 $\pm$ 0.75 \\
Multiple choice & SWAG & 81.08 $\pm$ 0.10 & 72.40 $\pm$ 0.09 & 63.80 $\pm$ 0.14 & 57.67 $\pm$ 0.16 \\
Fill-in-the-blank & Winogrande & 51.83 $\pm$ 1.11 & 51.49 $\pm$ 0.61 & 50.70 $\pm$ 0.81 & 50.01 $\pm$ 0.29 \\
Sentiment & IMDB & 93.54 $\pm$ 0.12 & 92.63 $\pm$ 0.06 & 89.09 $\pm$ 0.25 & 87.03 $\pm$ 0.23 \\
Comprehension & RACE & 66.25 $\pm$ 0.81 & 53.13 $\pm$ 1.10 & 48.35 $\pm$ 0.58 & 42.97 $\pm$ 0.32 \\
\bottomrule
\end{tabular}
\end{table*}

Two results need to be reported plainly rather than smoothed over.

\textbf{Winogrande is uninformative}, not evidence against SEWN specifically: every model, including full BERT-base, sits within roughly two points of chance (50\%). WinoGrande is explicitly constructed to resist the shortcuts a small training set (9{,}248 examples here) would otherwise let a model exploit~\cite{sakaguchi2020winogrande}; published results reaching well above chance use much larger training splits and larger backbones. We include the result for completeness, not as a claim about SEWN.

\textbf{RACE reveals a genuine capacity limitation that is not specific to routing.} DistilBERT drops 13.1 points relative to BERT-base here (53.13 vs.\ 66.25), a far larger relative collapse than its 0.3--8.7 point gaps on every other task, and SEWN's variants drop further still, with SEWN-sparse ($42.97 \pm 0.32$) landing 18 points above the 25\% random-choice floor but nonetheless the weakest result of any model on any task in this paper. Long-passage, four-way discriminative reading comprehension appears to require model depth/capacity that \emph{no} compressed model tested here retains, whether compressed via distillation or via routing. We cannot fully rule out that a fixed 3-epoch/$2\times10^{-5}$ recipe (tuned loosely for BoolQ) undertrains RACE specifically, but the fact that DistilBERT --- a properly, separately pretrained model --- collapses almost as badly argues against a SEWN-specific explanation.

Sentiment classification (IMDB) reproduces the BoolQ/PubMedQA pattern cleanly: DistilBERT stays close to BERT-base ($-0.91$ points), SEWN-sparse costs more ($-6.51$ points) but is $5.2\times$ faster (33.60ms to 6.42ms).

\subsection{Why is sparse routing's signal faithful? A falsification test}
\label{sec:falsification}

SEWN-sparse's content stream only ever sees the top-$k$ selected tokens --- tokens outside that selection have, by construction, zero direct path into the content stream's computation. This raises an obvious alternative explanation for Table~\ref{tab:pubmedqacf}'s strong faithfulness result: perhaps \emph{any} hard selection bottleneck produces strong top-vs-bottom separation, regardless of whether the selection criterion is any good, simply because excluded tokens structurally cannot matter to the prediction. We test this with three controls that isolate different parts of that hypothesis. SEWN-random-topk (Section~\ref{sec:randomtopk}) removes the ranking signal entirely (uniform random selection) to test whether the bottleneck alone is sufficient. SEWN-attn-topk adds a second, content-aware but non-learned ranking source --- the same attention-derived-significance idea used in the token-pruning literature (e.g.\ PoWER-BERT): tokens are ranked by a separate, frozen, already-fine-tuned BERT-base classifier's last-layer attention-to-CLS score, with no learned gate of SEWN's own involved at all. Both are architecturally identical to SEWN-sparse downstream of token selection; only the ranking source differs. The third control, Mixture-of-Depths~\cite{raposo2024mod}, is not a control we built to isolate one variable but an actual published routing method, reimplemented on a full 12-layer BERT-base-derived encoder (all weights transplanted from \texttt{bert-base-uncased}): every other layer (6 of 12) is ``routed'' --- a per-token linear router scores every position, the top 50\% by score are gathered and pass through that layer's full self-attention and FFN among themselves, the rest skip the layer via pure residual passthrough, with the router's contribution blended in through a sigmoid gate on the block's output so it receives gradient (the same differentiable-hard-selection trick SEWN-sparse's own gate already uses). This is a bidirectional-classification adaptation of a method published for causal language modeling; we state that adaptation plainly rather than let a reader discover it.

\begin{table}[t]
\centering
\caption{SEWN-sparse vs.\ three non-SEWN-gate ranking sources, PubMedQA, 3 seeds. SEWN-attn-topk's latency includes the frozen scorer's own forward pass. Mixture-of-Depths routes 6 of 12 layers at 50\% capacity (Section~\ref{sec:falsification}); its accuracy/latency are its own measured numbers, not SEWN's.}
\label{tab:randomtopk}
\begin{tabular}{lccccc}
\toprule
Variant & Accuracy & Latency & top-drop & bottom-drop \\
\midrule
SEWN-sparse & 82.66 $\pm$ 0.75 & 7.13ms & 0.328 & 0.006 \\
SEWN-random-topk & 80.09 $\pm$ 0.37 & --- & 0.043 & 0.031   \\
SEWN-attn-topk & 83.12 $\pm$ 0.18 & 64.60ms & 0.279 & 0.067  \\
Mixture-of-Depths & 83.24 $\pm$ 0.68 & 37.71ms & 0.124 & 0.052  \\
\bottomrule
\end{tabular}
\end{table}

SEWN-random-topk falsifies the ``bottleneck alone'' hypothesis outright: if hard selection alone were sufficient, its top-vs-bottom separation should be comparably strong --- it is not ($p = 0.45$, not significant), despite an identical mechanical guarantee that excluded tokens have no path into the content stream. \textbf{The learned ranking itself is doing the explanatory work}, not the architecture's exclusion mechanism in isolation.

SEWN-attn-topk sharpens this further by showing that \emph{some} non-learned rankings do carry real signal --- it passes the faithfulness test ($p=1.8\times10^{-13}$) and even edges out SEWN-sparse on raw accuracy (83.12 vs.\ 82.66) --- but on two dimensions it is still a strictly worse deal than SEWN's own learned gate. First, faithfulness: its bottom-drop (0.067) is an order of magnitude higher than SEWN-sparse's (0.006), meaning that the tokens it calls ``unimportant''still measurably matter to the prediction, unlike SEWN-sparse's near-total separation. Second, and more directly relevant to the compute-efficiency claim in Section~\ref{sec:sewnsparse}: producing this ranking requires a full extra frozen BERT-base forward pass, which makes SEWN-attn-topk's actual measured latency (64.60ms) not just slower than SEWN-sparse (7.13ms, a $9\times$ gap) but slower than plain, unpruned BERT-base itself (45.31ms, Table~\ref{tab:pubmedqa}). An attention-based heuristic ranking is not free as is sometimes implicitly treated in the pruning literature; SEWN's gate is cheap specifically because it is a small MLP computed once as part of the model's own forward pass, not a second model's worth of attention weights.

\begin{figure}[t]
\centering
\includegraphics[width=0.95\linewidth]{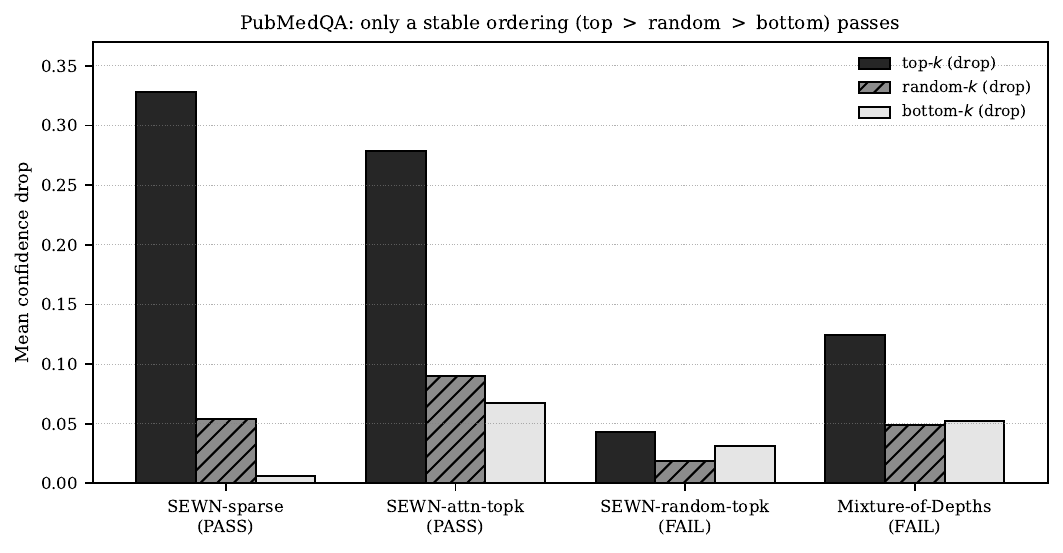}
\caption{Top/random/bottom-drop, all four Table~\ref{tab:randomtopk} methods. SEWN-random-topk's bars are jumbled (null result); Mixture-of-Depths clearly separates top from the other two but doesn't order bottom below random.}
\label{fig:modbars}
\end{figure}

The Mixture of depths is the more interesting of the two failing controls, and Table~\ref{tab:randomtopk}'s single $p$-value undersells why. Fig.~\ref{fig:modbars} plots top/random/bottom-drop for all four methods side by side: SEWN-random-topk's three bars are close together and out of order (a genuine null result, matching its $p=0.45$), while Mixture-of-Depths' top-drop (0.124) is clearly separated from \emph{both} its random-drop (0.049, $p=1.3\times10^{-6}$) and bottom-drop (0.052, $p=1.7\times10^{-4}$) --- both comparisons are highly significant. Under our updated strict ordering criterion ($\text{top} > \text{random} > \text{bottom}$), Mixture-of-Depths is classified as FAIL because its bottom-drop is nominally higher than its random-drop, even though its top-vs-random ($p = 1.3 \times 10^{-6}$) and top-vs-bottom ($p = 1.7 \times 10^{-4}$) separations are both highly significant. The MoD router reliably identifies key content tokens but fails to consistently separate bottom-ranked tokens from arbitrary ones. On the efficiency side, Mixture-of-Depths lands close to SEWN-attn-topk on accuracy (83.24 vs.\ 83.12) at less than half the wall-clock cost (37.71ms vs.\ 64.60ms), but its alternating-layer, 50\%-capacity routing is a much gentler compute cut than SEWN-sparse's upfront top-$k$ truncation, so its speedup over BERT-base is modest (1.20$\times$) next to SEWN-sparse's 6.36$\times$. The content-aware ranking signal is not sufficient by itself to match it on both faithfulness \emph{and} cost simultaneously (shown by SEWN-attn-topk), and a real published per-layer routing method with a genuinely informative router still does not reach the strict PASS bar SEWN-sparse clears (shown by Mixture-of-Depths) --- the learned gate's advantage is specifically that it produces a ranking this good \emph{for free} and this \emph{cleanly separated}, as a byproduct of the classification forward pass it was already going to run.

Two further observations sharpen the mechanistic account. First, the accuracy cost of random selection is modest (2.57 points) relative to its faithfulness cost ($7.7\times$ smaller top-drop) --- learned routing contributes disproportionately more to explanation reliability than to raw task accuracy. Second, we can explain \emph{why} random selection specifically fails the counterfactual test: the random selector redraws independently on every forward call. Masking the tokens it happened to select in one pass does not perturb the mechanism that produces the \emph{next} pass's selection, so a fresh, unrelated random draw at evaluation time often recovers equally useful tokens elsewhere in the sequence. The learned gate has no equivalent escape route: its score is a stable, content-determined function of each token's own embedding and local context, so masking the tokens it ranks highest removes evidence that a re-computation of the same gate, on the same (now-degraded) input, cannot replace. \textbf{Faithful attribution from a routing mechanism requires the selection criterion to be a stable function of content, not merely an information bottleneck.}

\subsection{Repairing a broken faithfulness result via distillation}
\label{sec:distill}

SEWN-lean (Section~\ref{sec:sewnlean}) removes the second stream, leaving the gate as the sole per-token mechanism with no stream to preserve a full-strength copy of whatever it suppresses.

\begin{table*}[t]
\centering
\caption{SEWN-lean, before and after distillation from a SEWN-contextual-gate teacher.}
\label{tab:distill}
\begin{tabular}{llcccccl}
\toprule
Variant & Task & Accuracy & top-drop & bottom-drop & $p$(top vs.\ bottom) & Verdict \\
\midrule
SEWN-lean & BoolQ & 68.33 $\pm$ 0.39 & 0.073 & 0.052 & 0.205 & \textbf{FAIL} \\
SEWN-lean & SWAG & 59.93 $\pm$ 0.03 & 0.330 & 0.033 & $9.2\times10^{-30}$ & PASS \\
SEWN-lean, distilled & BoolQ & 68.95 $\pm$ 0.26 & 0.099 & 0.022 & $1.6\times10^{-7}$ & \textbf{PASS} \\
SEWN-lean, distilled & SWAG & 60.76 $\pm$ 0.12 & 0.347 & 0.038 & $1.5\times10^{-33}$ & PASS \\
\bottomrule
\end{tabular}
\end{table*}

The KL-divergence logit plus an explicit MSE term matching the teacher's per-token gate score, $\mathcal{L} = \mathcal{L}_{\text{task}} + \alpha \cdot \mathcal{L}_{\text{logit-KD}} + \beta \cdot \mathcal{L}_{\text{gate-MSE}}$, $\alpha=1.0$, $\beta=2.0$) fixes BoolQ's broken faithfulness result outright ($p$: 0.205 to $1.6\times10^{-7}$) and closes most of the accuracy gap to the two-stream teacher (68.33 to 68.95, vs.\ teacher's 69.20), while SWAG's already-passing result stays intact and marginally sharpens. Distillation transferred a specific \emph{property} (a stable, content-determined gate ranking) from a teacher known to have it, rather than accuracy alone --- consistent with Section~\ref{sec:falsification}'s finding that faithfulness and accuracy are separable quantities that a single training objective does not automatically co-optimize.

\section{Discussion}

Putting Sections~\ref{sec:ablation}--\ref{sec:distill} together, four claims survive scrutiny and four do not.

\textbf{Survives:} (i) A hand-built linguistic prior can silently degrade a learned importance signal's reliability without any accuracy-metric symptom, on both backbones tested (Section~\ref{sec:staticvcontext}). (ii) Hard top-$k$ selection is a real, substantial compute win (5.2--8.7$\times$ over BERT-base, 2.6--4.4$\times$ over DistilBERT, across every dataset tested) that generalizes across a $3\times$ change in data scale and sequence length (Sections~\ref{sec:ksweep}--\ref{sec:pubmedqa}). (iii) A routing gate's faithfulness depends on the selection criterion being a stable, content-determined function --- not on the presence of a hard bottleneck per se (Section~\ref{sec:falsification}) --- and this specific property, not accuracy, is what distillation from a faithful teacher actually transfers (Section~\ref{sec:distill}). (iv) Against an actual published routing method (Mixture-of-Depths, reimplemented directly rather than only approximated by a heuristic control), SEWN-sparse's top-vs-bottom separation is still the cleanest in the comparison --- Mixture-of-Depths' router carries real signal (highly significant top-vs-random and top-vs-bottom separation) but does not clear the strict ordering PASS requires, while SEWN-sparse does (Section~\ref{sec:falsification}).

\textbf{Does not survive:} (i) That routing/gating meaningfully improves task accuracy over parameter-matched, non-adaptive baselines (Section~\ref{sec:ablation}). (ii) That a routing-based importance signal is something dense transformer baselines structurally cannot provide --- cheap post-hoc methods (raw attention) match it on faithfulness at near-zero cost (Section~\ref{sec:posthoc}). (iii) That the efficiency/faithfulness package generalizes across task types uniformly --- it holds for binary classification and sentiment, and collapses (for every compressed model tested, not only SEWN) on long-passage multiple-choice comprehension (Section~\ref{sec:taskbreadth}). (iv) That ``a contextual gate is more faithful than a static-prior gate'' is backbone-general --- it holds on BERT and on RoBERTa/SWAG, but on RoBERTa/BoolQ neither variant, at any prior size we tested, passes, and we could not identify a confound that explains the gap (Section~\ref{sec:staticvcontext}).

\textbf{Practical takeaway.} The honest, narrow claim this paper supports is: for long-context, single-judgment classification tasks, hard token-selection routing delivers a real additional speedup beyond what standard distillation already provides (2.6--4.4$\times$ over DistilBERT specifically, across every dataset tested), at a measured accuracy cost, bundled with an importance signal that is at least as reliable as the best cheap post-hoc alternative and considerably cheaper than the best expensive one. This is a narrower claim than ``SEWN is a better architecture,'' and we think that is the right size of claim for the evidence actually in hand.

\section{Limitations}
\label{sec:limitations}
\begin{itemize}
\item \textbf{No direct comparison against most of the token-pruning literature's actual released implementations} (PoWER-BERT, TR-BERT, Learned Token Pruning, ToMe). Section~\ref{sec:falsification} does include Mixture-of-Depths, reimplemented directly (not a heuristic proxy) on a full 12-layer BERT-base-derived encoder, and SEWN-attn-topk, a non-learned pruning baseline using the same attention-derived-significance idea several of these methods build on (ranking by a frozen classifier's attention-to-CLS score) --- together these let us report real accuracy/latency/faithfulness numbers against one genuine published routing method and one well-known heuristic, rather than only a qualitative comparison. Neither is a reimplementation of PoWER-BERT's progressive elimination schedule, TR-BERT's RL-learned policy, Learned Token Pruning's threshold mechanism, or ToMe's merge-based similarity ranking, and readers should not read Table~\ref{tab:randomtopk} as a substitute for benchmarking against those specific methods' released code. The Mixture-of-Depths reimplementation is also an adaptation (bidirectional classification, not the original's causal LM setting) rather than a reproduction of the published training procedure --- see Section~\ref{sec:falsification} for the adaptation stated explicitly.
\item \textbf{PubMedQA uses a custom, class-balanced subset of the \texttt{pqa\_artificial} configuration}, not the standard \texttt{pqa\_labeled} benchmark protocol used in most published PubMedQA results; absolute numbers are not comparable to the PubMedQA leaderboard.
\item \textbf{All experiments use BERT-base-scale backbones} (55--110M parameters). We make no claim about how these findings transfer to billion-parameter models, where the economics of routing-based compute savings and the baseline cost of post-hoc explanation methods could both look very different.
\item \textbf{The static-prior-vs-contextual-gate faithfulness result is backbone-dependent, not general.} Section~\ref{sec:staticvcontext} reports this plainly: on RoBERTa/BoolQ, the contextual gate configuration that passes cleanly on BERT ($p=1.7\times10^{-10}$) fails to reach significance ($p=0.177$), and we could not find a confound (prior size, coverage, or content, swept across the full 0--839 id range) that explains the gap. The result should be read as ``a large static linguistic prior never beats a contextual gate on the tasks tested,'' not as ``removing the prior reliably produces a faithful gate on any backbone.''
\item \textbf{RACE and Winogrande results are single-configuration} (one training recipe, not separately tuned per dataset); the RACE collapse in particular may partly reflect an undertuned recipe rather than a hard capacity ceiling, though the fact that a properly pretrained DistilBERT collapses nearly as much argues this is at most a partial explanation.
\item \textbf{Faithfulness evaluation is behavioral, not mechanistic}, in the sense used by Jain and Wallace~\cite{jain2019attention}: passing our counterfactual masking test establishes that masking the flagged tokens changes the prediction more than masking others, not that the flagged tokens are the unique or true causal explanation. Multiple valid rankings could in principle pass the same test.
\item \textbf{Faithfulness tests use $n=200$ examples per condition, and we report a large number of $p$-values across Tables~\ref{tab:staticvcontext}, \ref{tab:ksweep}, \ref{tab:pubmedqacf}, \ref{tab:posthoc}, \ref{tab:randomtopk}, and \ref{tab:distill} without a multiple-comparisons correction.} Most reported effects are large enough ($p < 10^{-10}$ in the majority of cases) that a standard correction (e.g.\ Bonferroni or Benjamini--Hochberg) would not change any PASS/FAIL verdict in this paper, but we did not apply one, and a smaller-$n$ replication with correction applied would strengthen these results further.
\end{itemize}

\section{Conclusion}

We set out to test, rather than assume, the two claims usually bundled into ``not all tokens need the same amount of attention.'' The efficiency claim holds, with real, scale-robust speedups, but at an accuracy cost that a reader needs to see reported honestly per task category, which we have tried to do. The interpretability claim is more fragile than it first appears: it depends on specific, testable architectural properties (a contextual rather than lexicon-anchored gate; a selection criterion that is a stable function of content, not merely a hard bottleneck) that are each falsifiable and, in one case in this paper, were falsified and had to be revised. We think the methodology here --- ablate the routing mechanism against parameter-matched non-adaptive baselines, test the importance signal against a real counterfactual protocol with random controls, and price any interpretability claim against the actual cost of the post-hoc alternatives it is implicitly being compared to --- generalizes beyond this specific architecture, and we would encourage other routing-based efficient-transformer work to apply the same tests before claiming an interpretability side-benefit.

\appendix
\section*{Appendix: Reproducibility Notes}

All experiments use a single NVIDIA RTX 4090, PyTorch 2.13, Transformers 5.14, seeds $\{42, 123, 456\}$, and the training recipe in Section IV held fixed across every variant and dataset unless explicitly noted as a swept parameter ($k$ for SEWN-sparse; the LR multiplier, validated once via a sanity sweep and then fixed). Result files (JSON) for every table in this draft are available in the accompanying code repository: \texttt{ablation\_results.json} (Table~\ref{tab:ablation}), \texttt{counterfactual\_results.json} / \texttt{counterfactual\_contextual\_gate\_results.json} (Table~\ref{tab:staticvcontext}), \texttt{sparse\_ksweep\_results.json} (Table~\ref{tab:ksweep}), \texttt{pubmedqa\_results.json} (Tables~\ref{tab:pubmedqa}--\ref{tab:pubmedqacf}), \texttt{explanation\_baseline\_results.json} (Table~\ref{tab:posthoc}, faithfulness columns), \texttt{task\_suite\_results.json} (Table~\ref{tab:taskbreadth}), \texttt{random\_topk\_ablation\_results.json} / \texttt{attn\_topk\_ablation\_results.json} (Table~\ref{tab:randomtopk}), \texttt{roberta\_gate\_check\_results.json} (Section~\ref{sec:staticvcontext}, backbone-generality check), \texttt{roberta\_prior\_confound\_results.json} / \texttt{roberta\_prior\_size\_sweep\_results.json} (Section~\ref{sec:staticvcontext}, prior size/coverage/content sweep, Fig.~\ref{fig:priorsize}), \texttt{mod\_pubmedqa\_results.json} (Table~\ref{tab:randomtopk} Mixture-of-Depths row, Fig.~\ref{fig:modbars}), \texttt{distill\_results.json} / \texttt{counterfactual\_lean\_results.json} / \texttt{counterfactual\_distilled\_lean\_results.json} (Table~\ref{tab:distill}).

\textbf{Exception:} Table~\ref{tab:posthoc}'s batched throughput/cost columns (\texttt{sewn\_throughput\_explanations.py}) were captured from console output only and have no saved JSON artifact --- re-run the script to reproduce those specific numbers before citing them in a camera-ready version.

\end{document}